\documentclass[runningheads]{llncs}
\usepackage{graphicx}
\usepackage{verbatim}
\usepackage{array}
\usepackage{multirow}
\usepackage{amsfonts}
\usepackage{stmaryrd}
\usepackage{amsmath}
\usepackage[caption=false]{subfig}

\usepackage{bbding}
\makeatletter
\newcommand{\printfnsymbol}[1]{%
  \textsuperscript{\@fnsymbol{#1}}%
}
\makeatother

\begin{document}
\title{Enhanced detection of fetal pose in 3D MRI by Deep Reinforcement Learning with physical structure priors on anatomy}

\titlerunning{Physical Structure Enhanced DRL on Fetal Pose}
%

%
\author{Molin Zhang\inst{1}\thanks{equal contribution}
\and
Junshen Xu\inst{1}\printfnsymbol{1} 
\and
Esra Abaci Turk\inst{2} \and
\\ P. Ellen Grant\inst{2,4} \and
Polina Golland\inst{1,3} \and
Elfar Adalsteinsson\inst{1,5}
}

\authorrunning{M. Zhang et al.}
%
\institute{Department of Electrical Engineering and Computer Science, MIT, \\ Cambridge, MA, USA \\\email{molin@mit.edu} \and
Fetal-Neonatal Neuroimaging and Developmental Science Center, \\ Boston Children’s Hospital, Boston, MA, USA \and
Computer Science and Artificial Intelligence Laboratory, MIT, \\
Cambridge, MA, USA \and
Harvard Medical School, Boston, MA, USA \and
Institute for Medical Engineering and Science, MIT, Cambridge, MA, USA}
%
\maketitle              

\begin{abstract}
Fetal MRI is heavily constrained by unpredictable and substantial fetal motion that causes image artifacts and limits the set of viable diagnostic image contrasts. Current mitigation of motion artifacts is predominantly performed by fast, single-shot MRI and retrospective motion correction. Estimation of fetal pose in real time during MRI stands to benefit prospective methods to detect and mitigate fetal motion artifacts where inferred fetal motion is combined with online slice prescription with low-latency decision making. Current developments of deep reinforcement learning (DRL), offer a novel approach for fetal landmarks detection. In this task 15 agents are deployed to detect 15 landmarks simultaneously by DRL. The optimization is challenging, and here we propose an improved DRL that incorporates priors on physical structure of the fetal body. First, we use graph communication layers to improve the communication among agents based on a graph where each node represents a fetal-body landmark. Further, additional reward based on the distance between agents and physical structures such as the fetal limbs is used to fully exploit physical structure. Evaluation of this method on a repository of 3-mm resolution in vivo data demonstrates a mean accuracy of landmark estimation within 10 mm of ground truth as 87.3\%, and a mean error of 6.9 mm. The proposed DRL for fetal pose landmark search demonstrates a potential clinical utility for online detection of fetal motion that guides real-time mitigation of motion artifacts as well as health diagnosis during MRI of the pregnant mother.

\end{abstract}
\section{Introduction}

Extracting localization of fetal pose plays a crucial role in fetal MRI~\cite{jokhi2011magnetic}. First, current fetal MRI is heavily constrained by the non-periodic and substantial fetal motion, thus substantial efforts are taken to mitigate motion artifacts by fast, single-shot MRI as well as retrospective methods like slice intersection registration~\cite{kim2009intersection}. Estimation of fetal pose during scanning may benefit prospective methods to detect and mitigate fetal motion artifacts via tracking of the motion and when combined with real-time, adaptive slice prescription. Further, fetal pose and fetal body movements may prove useful for monitoring fetal growth or other antenatal surveillance \cite{lai2016fetal,yen2019correlation}.

It's time consuming to manually annotate landmarks of fetal pose, and early demonstrations by Xu et al.~\cite{xu2019fetal} have addressed this problem with deep learning in convolution neural networks (CNN). Deep reinforcement learning (DRL)~\cite{mnih2015human} is a candidate for an alternative and powerful tool to handle this task. Previous work has been done to automatically detect single landmarks in medical imaging as a Markov Decision Process (MDP)~\cite{sutton1998introduction} and subsequently applying DRL \cite{ghesu2016artificial}. A multi-scale strategy~\cite{ghesu2017multi} and hierarchical action steps~\cite{alansary2019evaluating} have been proposed to further improve the performance of this approach. For the detection of multiple landmarks,~\cite{vlontzos2019multiple} proposed a collaborative DQN based on concurrent Partially Observable Markov Decision Process~\cite{girard2015concurrent}. However, the physical structure and communications among agents are not taken into consideration during decision making. Searching all landmarks of fetal pose, e.g., 15 agents as proposed for our case, is challenging due to the many degrees of freedom in fetal gesture and position, image artifacts and intensity variations. Yet, the fetal body is characterized by robust priors on physical connections between joints, and thus we expect that incorporating realistic skeletal priors will improve detection performance of key points for pose estimation. 

To exploit a prior on physical structure, the Graph Convolutional Network (GCN)
\cite{bruna2013spectral,defferrard2016convolutional,kipf2016semi} provides a powerful tool that represents each landmark of fetal pose as a node in a graph and combines features from both the center node and its neighbors. However, naive spatial domain GCN uses simple, fixed parameters in the adjacent matrix, which results in shared kernel weights over all edges where the relations of internal structure is not well exploited~\cite{kipf2016semi}. 
More recent efforts have been made to address this limitation by learning a semantic relationships among neighbor nodes implied in the edges~\cite{zhao2019semantic}. 

Moreover, most landmark-searching DRL only adopt an immediate reward based on the distance between the agent and its target landmark
\cite{alansary2019evaluating,ghesu2016artificial,ghesu2017multi,vlontzos2019multiple}. The smallest and most peripheral joints, i.e. ankles and wrists, represent the most challenging landmark detection task due to contrast and spatial resolution limits of the EPI acquisition, and to compound the difficulty of the problem, these joints tend to be the most mobile of the set of key points that characterize fetal pose. We expect that the incorporation of the physical structure of fetal anatomy will benefit the landmark identification and provide more robust features for these challenging joints~\cite{zhang2018deep}. Structure-aware regression on human pose estimation has been used by reparameterizing pose representation using bones instead of joints \cite{sun2017compositional,tang2018deeply}.

In the current work, we propose a novel end-to-end, multi-agent deep reinforcement learning network to detect fifteen landmarks of fetal pose in each frame of a time series of volumetric fetal MRI. Fifteen independent observed MRI volumes are used as input through a parameter shared convolutional network generating fifteen hidden feature presentations. Then fifteen dense layers with graph communication layers process the corresponding fifteen features for better action decisions. Further, an additional and immediate reward is designed based on the distance between the agents and physical structure connections, such as limbs that could help modify the optimal search path through a more robust structure and to improve performance. Our method achieved 87.25\% accuracy and 6.9 mm as the mean error of landmark detection.  

\section{Methods}
\subsection{Deep reinforcement learning for landmark detection}
The task of landmark identification in fetal MRI fits a conventional MDP reinforcement learning framework. We identify five MDP components, $\mathcal{M} = (\mathcal{S}, \mathcal{A}, \\ \mathcal{R},\mathcal{T},  \gamma)$. $\mathcal{S}$ represents the set of states. $s_t^k \in \mathcal{S}$ is the state at step $t$ of agent $k$, that is a $48\times48\times48$ MRI volume centered at the position of corresponded searching agent. $\mathcal{A}$ represents the set of actions, namely, moving \textit{forward}, \textit{backward}, \textit{left}, \textit{right}, \textit{upward} and \textit{downward}. $\mathcal{R}$ represents the immediate reward based on $s_t^k$ and $a_t^k$. $\mathcal{T}$ is the transition function describing the sequential state under $s_t^k$ and $a_t^k$, which is deterministic in our case. $\gamma$ is the discount factor balancing immediate reward and future reward.

Due to the large number of dimension of $\mathcal{S}$, a single-agent RL samples states from MDP and optimizes the target function iteratively by using the experience. There are several algorithms to solve the optimization process. One of them is Q-learning~\cite{watkins1992q} in which we consider the following state-action function
\begin{equation}
    Q(s,a) = \mathbb{E}\left[\sum_{i=1}^n\gamma^{i-1} r_{t+i} | s, a\right]
\end{equation}

Q-learning learns the optimal action policy by finding the highest expected future return, annotated as $Q^{*}(s, a)$. With the Bellman equation, it could be rewritten recursively as  $Q^*(s_{t}, a_{t})=E\left[r_t+\gamma \max_{a_{t+1}} Q^*\left(s_{t+1}, a_{t+1}\right)\right]$ at step $t$. 
With the development of deep neural networks, DQN is proposed to approximate $Q(s, a ; \theta) \approx Q^{*}(s, a)$ using a neural network~\cite{mnih2015human}. 

\subsection{Multi-agent RL with a graph communication layer}
 In terms of multi-agent RL for landmark detection, collab-DQN~\cite{vlontzos2019multiple} adopts a shared CNN to extract features from the observed environment of each agent, then estimates $Q(a,s)$ for each agent independently using separate, fully connected networks. This architecture, however, fails to share information across different agents during decision making, which is important for detecting fetal landmarks that are spatially correlated.

\begin{figure}[ht]
\centering
\includegraphics[width=0.9\textwidth]{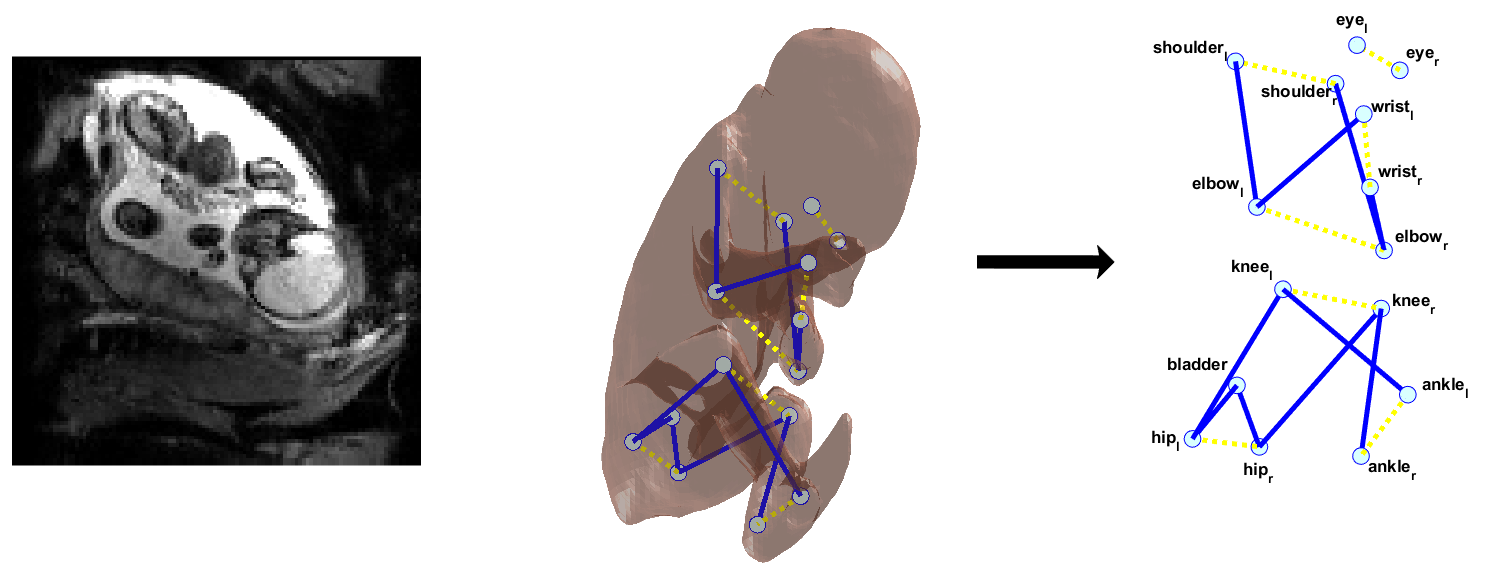}
\caption{\textbf{Left}: One slice of a 3D BOLD EPI volume. \textbf{Middle and Right}: Contour of fetus and physical structure graph based on pose landmarks. The nodes are 15 landmarks including eyeballs, bladder and joints which are annotated in the figure. There are two types of edge. The blue solid line represents edges of skeleton connection and the yellow dashed line represents edges of right and left landmark correlation.} \label{fig:data}
\end{figure}

Given a graph with $K$ nodes, Graph convolution~\cite{kipf2016semi} enables feature extraction from graph-structured data by combining the feature of each node with features of its neighbors, i.e., $\mathbf{H}=\sigma(\mathbf{W}\mathbf{X}\mathbf{\tilde{A}})$, where $\sigma$ is the activation function,  $\mathbf{H}\in\mathbb{R}^{C_o\times K}$ and $\mathbf{X}\in\mathbb{R}^{C_i\times K}$ are output and input features, $\mathbf{W}\in\mathbb{R}^{C_o\times C_i}$ is a learnable parameter matrix, and $\mathbf{\tilde{A}}\in\mathbb{R}^{K\times K}$ is symmetrically normalized from adjacency matrix $\mathbf{A}$~\cite{biggs1993algebraic}. However, conventional GCN uses fixed parameters in the adjacent matrix which cannot model the difference in the contribution of neighbors to the target node. One way to address this problem is to make adjacent matrix trainable so that the network can learn a semantic relationships of neighbor nodes implied in the edges~\cite{zhao2019semantic}. Another limitation of GCN is the shared kernel matrix $\mathbf{W}$ which constrains the representation ability where the fact is the correlation between each node in physical structure graph is not equal. We propose the following graph communication layer, 

\begin{equation}
    \mathbf{H}=\sigma\left( \bigparallel_{i=1}^N\left(\mathbf{W}^{(i)}x_i\right) \rho(\mathbf{M} \odot \mathbf{A})\right)
    \label{semGcn}
\end{equation}
where $\mathbf{W}^{(i)}$ and $x_i$ are the kernel matrix and input feature of node $i$, $\bigparallel$ represents concatenation, $\rho$ is row-wise softmax, $\mathbf{M}$ is a trainable matrix, and $\odot$ represents element-wise operation defined in~\cite{zhao2019semantic}, resulting in a trainable adjacent matrix.

In this work, as shown in Fig. \ref{fig:data}, we choose 15 landmarks (eyeballs, bladder, shoulders, elbows, wrists, hips, knees, and ankles) as the representation of fetal pose, as in~\cite{xu2019fetal}. Then we build the graph where each node represents a landmark of fetal pose and two types of edges are considered. One is the physical connection of fetal skeleton and the other is the connection of the left- and right-side landmarks.

\begin{figure}[ht]
\centering
\includegraphics[width=0.9\textwidth]{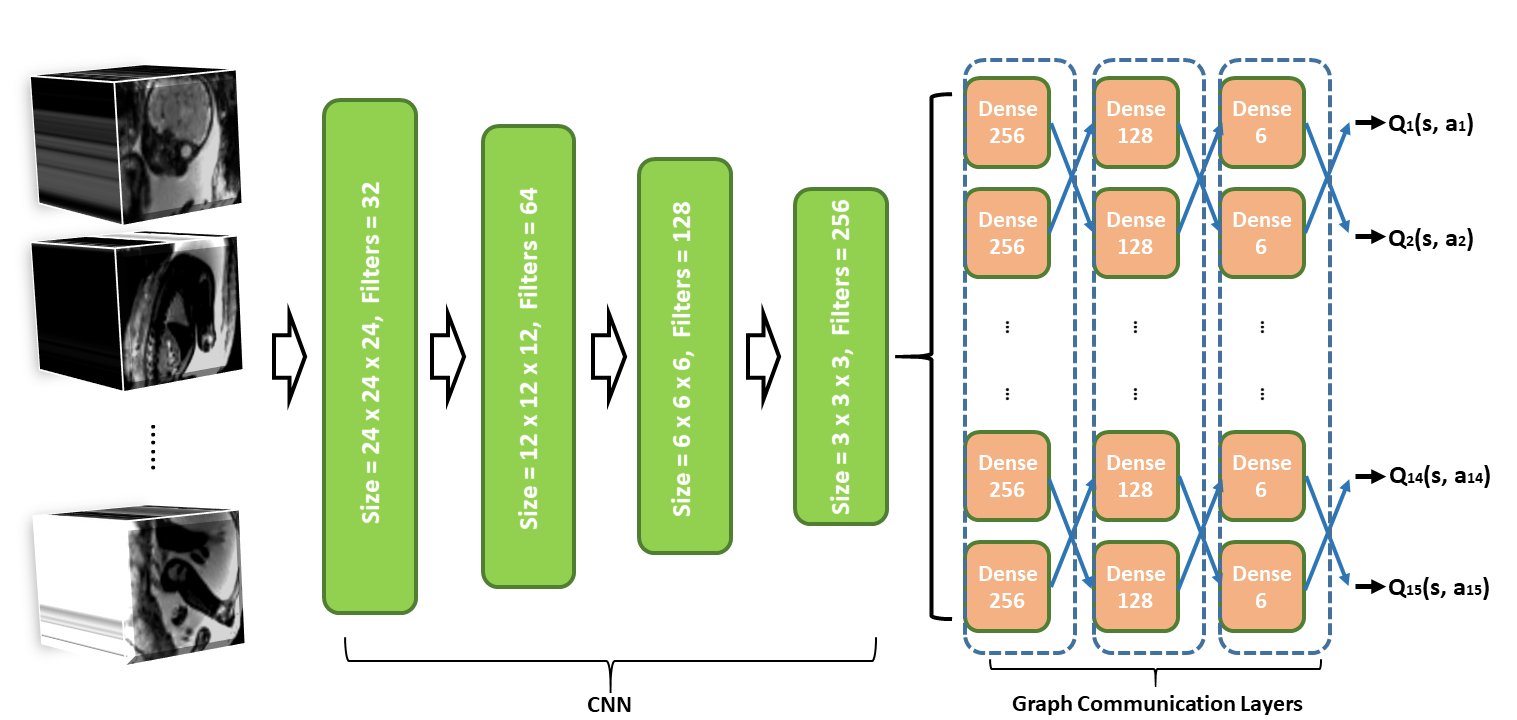}
\caption{The architecture of our deep reinforcement learning framework. The green box represents the convolutional block which consists two stacks, each with one 3D convolutional layer with kernel size of 3, stride step of 2, one batch-normalization layers, and one activation layers using ReLU. There are four convolutional blocks with pooling layers of stride 2 following. The channels of each blocks are 32, 64, 128, and 256. The blue dashed box represents graph communication layers (Equation \ref{semGcn}) where there are three graph communication layers with hidden nodes equals to 256, 128, 6.} \label{fig:architecture}
\end{figure}

The proposed architecture for multi-agent landmark detection is illustrated in Fig. \ref{fig:architecture}. First, a shared convolution network is used to extract features from the observed environment of each agent. Then the graph-structured features are fed into a three-layer graph communication network to merge the information of correlated landmarks before producing the final estimation of $Q$ functions. More network details are listed in the caption. The loss function is defined as,
\begin{equation}
        L(\theta)=\sum_{k=1}^{15}\mathbb{E}\left[\left(r^k_t+\gamma \max _{a^k_{t+1}} Q_k\left(s_{t+1}, a^k_{t+1} ; \theta^{-}\right)-Q_k(s_t, a^k_t ; \theta)\right)^{2}\right]
     \label{eq:dqn}
\end{equation}
where $i$ is the index of each agent, $\theta$ is parameters of the target deep neural network and $\theta^{-}$ is parameters of a frozen network which will be updated by the target network every fixed iterations~\cite{mnih2015human}. 

\subsection{Physical structure reward}
Most DRL in medical landmark searching use immediate reward design based on the distance between the location of the agent and corresponding landmark. For fetal MRI with low-resolution EPI imaging, landmarks like wrists, ankles and elbows have large location distribution and shape variations, but fetal limbs are relatively stable and present with robust signal intensity on these scans and provide a strong and clear physical connection between joint landmarks. We propose an elaboration of the reward design by incorporating the distance between agent and the corresponding limbs such that the reward is used for agent sets $\mathcal{S}_l$ (\textit{shoulders, elbows, wrists, hips, knees and ankles}) searching landmarks on limbs. The reward for agent $k$ at time step $t$ is defined as below,
\begin{equation}
    r_t^k = D_l(t,k) - D_l(t+1,k) + \beta\sum_{m\in\mathcal{N}(k)}\left(D_b(t,k,m) - D_b(t+1,k,m)\right)
    \label{reward}
\end{equation}
where $\mathcal{N}(k)$ is the neighbor of agent $k$ on the corresponded limb. $D_l(t,k) = \left\|p_{k,t}-p^{GT}_{k}\right\|_{2}$ is the Euclidean distance between the location of agent $p_{k,t}$ at step $t$ and corresponded landmark $p^{GT}_{k}$. $D_b(t,k,m)$ is the the Euclidean distance between the location of agent and the segment connecting landmark $k$ and $m$.
\begin{equation}
D_b(t,k,m)=\left\{\begin{aligned}
&\left\|p_{k,t}-p^{GT}_{k}\right\|_{2},\text{\quad\quad if }(p_{k,t} - p^{GT}_k)\cdot a_{km} > 0\\
&\left\|p_{k,t}-p^{GT}_{m}\right\|_{2},\text{\quad\quad if }(p_{k,t} - p^{GT}_m)\cdot a_{km} < 0\\
&\left\|(p_{k,t} - p^{GT}_m)\times a_{km}\right\|_{2}/\left\|a_{km}\right\|_{2},\text{ otherwise}
\end{aligned}\right.
\end{equation}
where $a_{k,m} = p^{GT}_k- p^{GT}_m$.

$\beta$ is a parameter that needs to be adjusted to achieve the desired balance during search, where a larger $\beta$ will force the agent to gravitate more aggressively towards limbs. This will help the CNN to learn more about the features of physical structures, as the optimal path would cover larger limb area, but it will also increase the difficulty of training convergence as the relative effect of reward on moving towards target landmarks is weakened.

\section{Experiments and Results}
\subsection{Dataset}
The dataset consists 19,816 3D BOLD MRI volumes acquired on a 3T Skyra scanner (Siemens  Healthcare,Erlangen,  Germany) with multislice, single-shot, gradient echo EPI sequence. The in-plane resolution is $3\times3mm^2$ and slice thickness is $3mm$. The FOV captures the whole uterus and thus has a variable matrix size from subject to subject. Fig. \ref{fig:data} shows one slice of an MRI volume. The gestational age range of the 70 fetuses ranged from 25 to 35 weeks. The mean matrix size was $120\times120\times80$; TR=$5-8$s, TE=$32-38$ms, FA=90$^{\circ}$. 

All 70 fetuses, 19,816 MRI volumes were manually labeled; 49 fetuses, 12,332 volumes were used for training; 14 fetuses, 3,402 volumes were used for validation; 14 fetuses, 4,082 volumes were used for testing.

One bottleneck of DRL is the generalization. To mitigate this, all training datasets were randomly flipped and rotated, which increases the variation of spatial distribution and, further, were randomly scaled with factor in the range of 0.8 $\sim$ 1.5. 

\subsection{Experiments setup}
During training, we used distributed DRL in multiple GPUs with DQN as the optimization algorithm.  All experiments were performed on a server with an Intel Xeon E5-1650 CPU, 128GB RAM and four NVIDIA TITAN X GPU.

Based on~\cite{horgan2018distributed}, our distributed DRL framework consists of $N$ actors and $M$ learners. Each actor fetches parameters from the global model periodically and generates experiences following the current policy, which are sent to shared memory. In each training step, each learner will copy the global model as well as a batch of experiences from shared memory. It then computes the local gradient of the DQN loss function (equation \ref{eq:dqn}) and pushes it to a global optimizer. We use Adam as the global optimizer which collects gradients from learners and updates the global asynchronously to avoid congestion due to lock.

In the experiments, We use $N=4$, $M=4$, batch size of 3, action step of 1 and learning rate of $3\times10^{-4}$. 50 is the maximum threshold for gradient clipping. 

\subsection{Results}
In this section, We evaluate the performance of our proposed enhanced DRL incorporating physical structure prior. We compare it with (1) conventional DRL whose frame work and architecture is similar to previous work~\cite{vlontzos2019multiple}, (2) DRL with graph communication layers, abbr. DRL+GC,  (3) DRL with physical structure reward, abbr. DRL+PR, (4) deep learning method with Unet~\cite{xu2019fetal}.

For fair comparison, all DRL methods are trained under same protocol described in Section. 3.2. $\beta = 2$ is chose for the physical reward weight. The DL method is trained using the same encoder layers as the shared convolutional network in DRL. The size of the input maintains $48^3$ voxels for all experiments.

The evaluation metrics are defined by a) Percentage of Correct Keypoint (PCK), the ratio of detected points if the distance between the detected and the true landmark is within a certain threshold, and b) mean error (in  mm), i.e., the mean distance between detected and ground-truth landmarks. The initial location of landmark agents is randomly distributed inner 15\% of the volume for both training and evaluation. All DRL evaluation experiments are repeated three times. On average, it takes 1s/volume during inference process.

\begin{table}[ht]
\caption{PCK (10mm) performance and mean error of different models.}
\label{tab1}
\begin{center}
\begin{tabular}{c|c|ccccccccc}
\hline
metric &method & eye &  shoulder& elbow  & wrist & bladder & hip & knee & ankle & all\\
\hline
\multirow{5}{*}{\shortstack{PCK\\(\%)}}
&Unet  &90.8 &   87.51   &  75.18          &   50.99        &92.99        &63.63   &87.08     &\textbf{56.01}     &74.37  \\
&DRL              &94.60               &96.81            &78.50  &45.79           &96.42 &74.86    &80.56    &31.46      &73.44  \\
&DRL+GC    &91.33     &92.19               &86.70   &68.10    &97.04          &83.38       &92.33     &42.93     & 80.73 \\
&DRL+PR  &94.56     &97.93      &81.34  &55.23    &96.70          &86.41       &86.42     &35.54     &78.10\\
&Proposed  &\textbf{97.35}     &\textbf{99.79}      &\textbf{91.21}  &\textbf{74.29}    &\textbf{98.33}         &\textbf{92.27}         &\textbf{94.75}    &55.58     &\textbf{87.25}\\
\hline
\multirow{5}{*}{\shortstack{mean\\(mm)}}
&Unet  &6.33& 11.61&8.29&24.87&6.99&25.74&9.60&23.67&18.63\\
&DRL&5.16 & 5.64 & 12.30 & 34.08 & 9.21 & 18.99 & 10.05 & 35.73 & 16.86\\
&DRL+GC & 5.43& 7.56& 6.99& 14.37 & 5.73 & 8.19& 6.00 & 23.13 & 9.93\\
&DRL+PR  &4.14& 3.60& 10.50&17.49& 6.12& 8.97&8.49& 28.53& 11.31\\
& Proposed &\textbf{2.94} & \textbf{2.37} & \textbf{5.70} & \textbf{10.74} & \textbf{3.93} & \textbf{4.20} & \textbf{4.26} & \textbf{19.53} & \textbf{6.90}\\
\hline
\end{tabular}
\end{center}
\end{table}

Table \ref{tab1} shows the results of all five experiments. Our proposed physical structure enhanced DRL achieves a significant better average detection accuracy under 10mm at 87.3\% and outperforms other models in most landmarks. Also, the mean detected error of all landmarks is 6.9 mm which is the smallest. Compared with Unet, the worse PCK accuracy but smaller mean error on ankle shows the fact that DRL agent detected the target ankle landmark but scattering around it due to the large size of ankle and foot, while Unet has more false detected cases. To some extent, our proposed method still behaves well.

\begin{figure}[ht]
\centering
\subfloat[\label{fig:hist}]{%
  \includegraphics[width=0.35\textwidth]{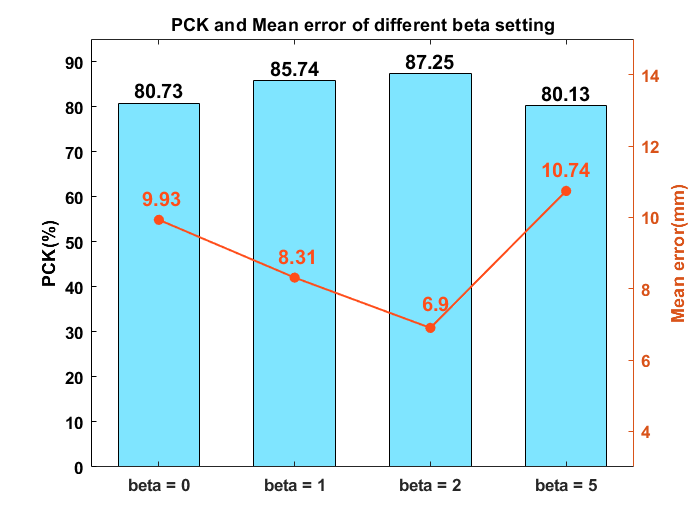}%
}\hfil
\subfloat[\label{fig:path2}]{%
  \includegraphics[width=0.65\textwidth]{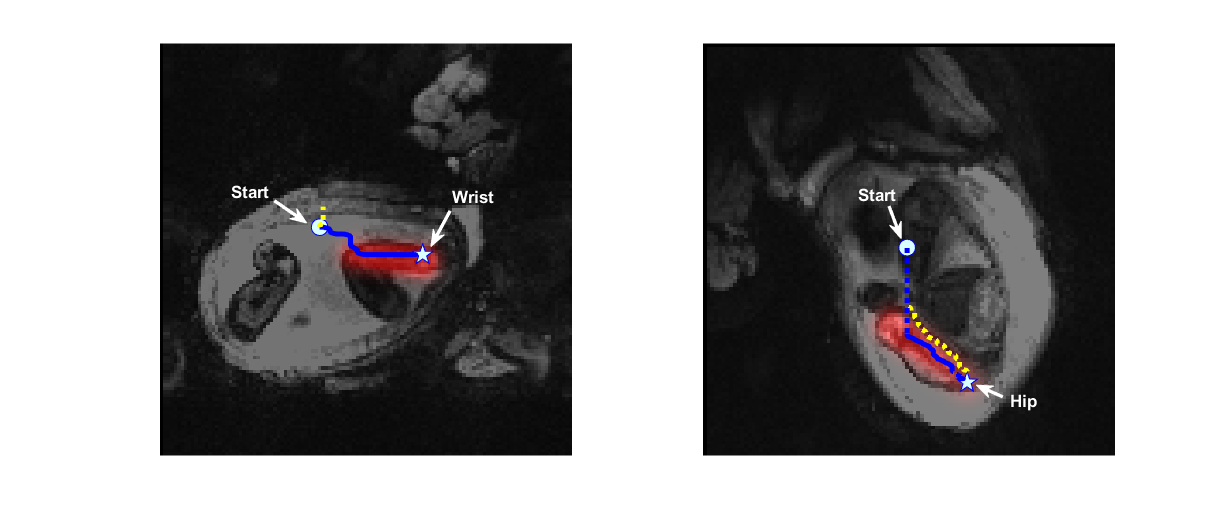}%
}
\caption{(a) PCK and mean error performance of different $\beta$ settings. (b) Illustration of searching paths of DRL with and without addition physical reward(PR). Blue line is the path of DRL w/ PR and yellow lines is the path of DRL w/o PR. Solid lines represent paths in current slice and dashed lines represent path in other slices. Red area represents corresponded limb in PR. The left-hand side image shows the successful case where DRL w/ PR recognizes limb structure and goes through it while DRL w/o PR fails. The right-hand side image shows the modified searching path of limb-based reward which will gravitate the agent towards the limb while another will go straight instead.}
\label{fig:path}
\end{figure}

In an ablation study, we compared the effects of different weights of physical reward in equation \ref{reward}. We choose $\beta = 0, 1, 2, 5$ with graph communication layers. Seen from Fig. \ref{fig:hist}, the model achieves best performances when $\beta =2$. Consistent with the effect of physical reward, when $\beta$ is too small, e.g. $\beta = 0$, the network intends to lack of feature extraction ability of physical structure. When $\beta$ is too large, e.g. $\beta = 5$, the agent will get trapped in extracting physical structure features and moving closer instead of moving towards target landmark.

Fig. \ref{fig:path2} shows the effect of physical reward (PR). The blue and yellow paths represents DRL with and without PR respectively. On the left, both methods use the same initial location while the distance between the agent and target landmark, right wrist, is over 30 voxels which is outside the FOV of the initial location. The agent of DRL without PR converges to the wrong location, both because the target landmark is beyond the 24-voxel distance, as well as weak feature analysis ability of the surrounding structure. The agent of DRL with PR succeeds by observing the structure of the arm. The right-hand side image shows the effect of modifying searching path where the agent is likely to go through robust structures like limbs which improves accuracy of detection.

\section{Conclusion}
In this work we proposed the incorporation of physical fetal skeletal structure to enhance DRL to infer landmarks of fetal pose in low-resolution 3D MRI in pregnancy. The proposed method achieves an average detection accuracy of 87.3\% under a 10-mm threshold and 6.9 mm as the mean error.

Our proposed approach exploits graph communication layers that benefit the communication among agents, as well as rewards derived from proximity to skeletal fetal structure that enhances network performance where search paths gravitate towards skeletal voxels as they approach landmark joints. The significantly convincing performance indicates the success of our proposed method. 

Overall, this work demonstrates the potential of using our proposed method to rapidly and robustly detect fetal landmarks for clinical utility that includes estimation of fetal pose, monitoring of fetal motion in health and disease, and motion tracking for real-time slice prescription to mitigating motion artifacts in diagnostic MRI in pregnancy. 

\section*{Acknowledgements}
This research was supported by NIH U01HD087211, NIH R01EB01733 and NIH NIBIB NAC P41EB015902.

\bibliographystyle{splncs04}
\bibliography{reference}

\end{document}